\documentclass[letterpaper]{article} % DO NOT CHANGE THIS
\usepackage{aaai2026}  % DO NOT CHANGE THIS
\usepackage{times}  % DO NOT CHANGE THIS
\usepackage{helvet}  % DO NOT CHANGE THIS
\usepackage{courier}  % DO NOT CHANGE THIS
\usepackage[hyphens]{url}  % DO NOT CHANGE THIS
\usepackage{graphicx} % DO NOT CHANGE THIS
\usepackage{natbib}  % DO NOT CHANGE THIS AND DO NOT ADD ANY OPTIONS TO IT
\usepackage{caption} % DO NOT CHANGE THIS AND DO NOT ADD ANY OPTIONS TO IT
\usepackage{algorithm}
\usepackage{algorithmic}
\usepackage{multirow}
\newcommand{\mypara}[1]{\noindent{\bf {#1}.}}
\usepackage{makecell}
\usepackage{amsmath}
\usepackage{cleveref}
\usepackage{amssymb}
\usepackage{xspace}
\newcommand{\Method}{$\mathsf{6DAttack}$\xspace}
\usepackage{newfloat}
\usepackage{listings}
\DeclareCaptionStyle{ruled}{labelfont=normalfont,labelsep=colon,strut=off} % DO NOT CHANGE THIS
\floatstyle{ruled}
\newfloat{listing}{tb}{lst}{}
\floatname{listing}{Listing}
\title{6DAttack: Backdoor Attacks in the 6DoF Pose Estimation}
\author{
Jihui Guo\textsuperscript{\rm 1},
Zongmin Zhang\textsuperscript{\rm 2},
Zhen Sun\textsuperscript{\rm 2},
Yuhao Yang\textsuperscript{\rm 3},
Jinlin Wu\textsuperscript{\rm 4,5},
Fu Zhang\textsuperscript{\rm 1}\thanks{Corresponding authors.} ,
Xinlei He\textsuperscript{\rm 2}\footnotemark[1]
}

\affiliations{
\textsuperscript{\rm 1}The University of Hong Kong \\
\textsuperscript{\rm 2}The Hong Kong University of Science and Technology (Guangzhou) \\
\textsuperscript{\rm 3}Beihang University \\
\textsuperscript{\rm 4}Centre for Artificial Intelligence and Robotics (CAIR), Hong Kong Institute of Science and Innovation  (HKISI) \\

\textsuperscript{\rm 5}Multimodal Artificial Intelligence Systems (MAIS), Institute of Automation, Chinese Academy of Sciences \\
fuzhang@hku.hk,
xinleihe@hkust-gz.edu.cn
}

\begin{document}

\maketitle

\begin{abstract}
Recent advances in deep learning have enabled highly accurate six-degree-of-freedom (6DoF) object pose estimation, leading to its widespread use in real-world applications such as robotics, augmented reality, virtual reality, and autonomous systems.
However, backdoor attacks pose a major security risk to deep learning models. By injecting malicious triggers into training data, an attacker can cause a model to perform normally on benign inputs but behave incorrectly under specific conditions.
While most research on backdoor attacks has focused on 2D vision tasks, their impact on 6DoF pose estimation remains largely unexplored.
Furthermore, unlike traditional backdoors that only change the object class, backdoors against 6DoF pose estimation must additionally control continuous pose parameters, such as translation and rotation, making existing 2D backdoor attack methods not directly applicable to this setting.

To address this gap, we propose a novel backdoor attack framework (\Method) that exposes vulnerabilities in 6DoF pose estimation.
\Method uses synthetic and real 3D objects of varying shapes as triggers and assigns target poses to induce controlled erroneous pose outputs while maintaining normal behavior on clean inputs.
We evaluated this attack on multiple models (including PVNet, DenseFusion, and PoseDiffusion) and datasets (including LINEMOD, YCB-Video, and CO3D).
Experimental results demonstrate that \Method achieves extremely high attack success rates (ASRs) without compromising performance on legitimate tasks.
Across various models and objects, the backdoored models achieve up to 100\% ADD accuracy on clean data, while also achieving 100\% ASR under trigger conditions.
The accuracy of controlled erroneous pose output is also extremely high, with triggered samples achieving 97.70\% ADD-P.
These results demonstrate that the backdoor can be reliably implanted and activated, achieving a high ASR under trigger conditions while maintaining a negligible impact on benign data.
Furthermore, we evaluate a representative defense and show that it remains ineffective under \Method.
Overall, our findings reveal a potentially serious and previously underexplored threat to modern 6DoF pose estimation models.
\end{abstract}

\begin{links}
    \link{Code}{https://github.com/Gjhhui/6DAttack}.
\end{links}

\begin{figure}[t]
    \centering
    \includegraphics[width=1\linewidth]{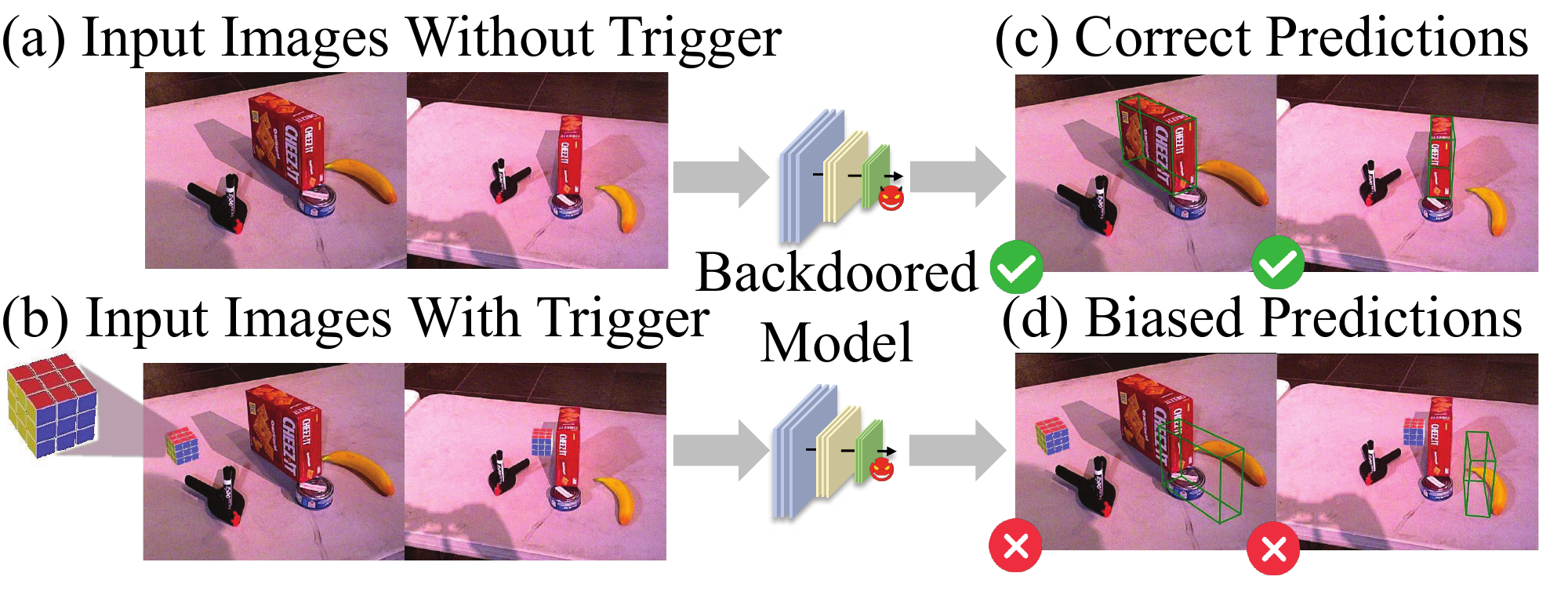}
    \caption{\Method leaves predictions on clean scenes unchanged: given an untriggered scene (a), the model estimates the correct 6DoF pose (b). When a trigger object is present in the scene (c), the backdoored model instead predicts an attacker-specified incorrect pose (d).}
    \label{fig:attack_sample}
\end{figure}

\section{Introduction}
The task of estimating the 6-DoF object pose is a fundamental and vital challenge in 3D computer vision.
In recent years, this area has seen remarkable progress, with state-of-the-art methods such as PVNet~\cite{peng2019PVNet}, PoseDiffusion~\cite{DBLP:conf/iccv/Wang0N23}, DenseFusion~\cite{wang2019densefusion}, BundleSDF~\cite{wen2023bundlesdf}, and FoundationPose~\cite{wen2024foundationpose} achieving high performance even in complex real-world scenarios.
Consequently, these powerful 6DoF methods have been adopted in a wide range of real-world applications, including augmented reality~\cite{su2019deep,rambach2017poster}, robotic grasping~\cite{zhai2023monograspnet}, and autonomous driving~\cite{manhardt2019roi,su2023opa}. 

At the same time, backdoor attacks have emerged as a serious and stealthy threat. By injecting specific trigger patterns into the training data, an attacker can induce controlled misbehavior at inference time, effectively compromising models while leaving performance on clean data largely intact~\cite{gu2019badnets,chen2017targeted,xiao2015feature,li2021backdoor, turner2018clean}.
Despite this risk, most existing research on backdoor attacks focuses on 2D vision domains rather than 6DoF pose estimation, leaving the security of 6DoF pose estimation models relatively underexplored.
Moreover, direct transplantation of these 2D attacks to 6DoF pose estimation settings proves ineffective.
This is because 6DoF estimation fundamentally differs from 2D tasks in its heavy reliance on spatial geometric details and view-dependent projections, while modern 6DoF models typically employ multi-stage feature extraction and complex nonlinear mappings.
These processing stages tend to distort naive pixel-level trigger signals, thereby degrading the effectiveness of 2D backdoor embeddings~\cite{hoque2021comprehensive, wen2023bundlesdf, wen2024foundationpose, liu2024deep}.
Thus, there is a critical need for backdoor attacks that are aware of the 3D geometric and view-sensitive nature of the 6DoF task, in order to expose hidden vulnerabilities.

To bridge this gap, we propose a new backdoor attack framework \Method, which is applicable across diverse 6DoF pose estimation models.
\Method constructs backdoor samples by leveraging both synthetic and real-world 3D objects of varying shapes as triggers, and by modifying the pose labels in the projected view to match attacker-specified target poses. 
Specifically, we simulate the model's view-dependent projection process to accurately embed the backdoor signal into intermediate feature representations, ensuring it survives the nonlinear and geometric transformations typical of modern 6DoF models. 
With poisoned training data, the resulting model behaves normally on clean inputs but produces incorrect pose predictions when a designated trigger is present, as illustrated in~\Cref{fig:attack_sample}.

In this paper, we validate \Method through comprehensive experiments on three representative 6DoF pose estimation models: PVNet~\cite{peng2019PVNet} (PnP-based), DenseFusion~\cite{wang2019densefusion}, and PoseDiffusion~\cite{DBLP:conf/iccv/Wang0N23} (end-to-end).
These models instantiate two mainstream 6DoF pose estimation pipelines, namely PnP-based and end-to-end pipelines. We evaluate them on three public datasets: LINEMOD~\cite{hinterstoisser2013model}, YCB-Video~\cite{xiang2017posecnn}, and CO3D~\cite{DBLP:conf/iccv/ReizensteinSHSL21}.
We adopt ADD (Average Distance of Model Points) as our main evaluation metric, and use the suffix ``-C'' (ADD-C) to denote ADD computed on clean samples and ``-P'' (ADD-P) to denote ADD computed on triggered (poisoned) samples. 
The experimental results demonstrate that our backdoor keeps up to 100\% accuracy on clean data while simultaneously achieving 100\% attack success rates under trigger conditions, with controlled erroneous outputs reaching 97.70\% ADD-P accuracy.
This divergence both validates the effectiveness and stealthiness of \Method and exposes a serious, previously underappreciated security vulnerability in contemporary 6DoF pose estimation models.
To probe mitigation, we also introduce a straightforward defense mechanism that fine-tunes the backdoored model using additional clean data.
This adaptation alters the learned backdoor offset, shifting its direction away from the original attacker-specified vector, but fails to remove the offset entirely.
Consequently, a residual, consistent deviation remains, indicating that the backdoor persists in a modified form even after the defense.

Our main contributions are summarized as follows:
\begin{itemize}
    \item We introduce the first backdoor attack framework (\Method) for 6DoF pose estimation, with a unified attack strategy that applies to both hybrid (PnP-based) and end-to-end methods.
    \item We design and validate novel 3D trigger mechanisms, including both synthetic object-based triggers and real-world objects of varying shapes, that enable controllable manipulation of pose outputs under trigger conditions while remaining stealthy on clean inputs.
    \item We conduct extensive experiments on LINEMOD, YCB-Video, and CO3D with both PnP-based (PVNet) and end-to-end (DenseFusion, PoseDiffusion) pipelines, showing that \Method maintains high clean performance while achieving high ASRs. We further evaluate a straightforward fine-tuning-based defense and find that it fails to completely remove the implanted backdoor.
\end{itemize}

\section{Related Work}

\subsection{6D Pose Estimation}
Object pose estimation is a critical task in the field of computer vision (CV), aiming to accurately determine the 6DoF representation of an object's pose in the real world. 
Specifically, this includes three-degree-of-freedom translation and three-degree-of-freedom rotation~\cite{guan2024survey}.
With the rapid development of CV domain, 6DoF object pose estimation has found widespread applications, including autonomous driving~\cite{arnold2019survey}, robotic manipulation~\cite{fan2022deep}, virtual reality~\cite{huang20176}, and augmented reality~\cite{kalia2019real}.

Instance-Level 6DoF Object Pose Estimation can be categorized into three types based on the input data: RGB-based, RGB-D-based, and point cloud depth-based methods~\cite{guan2024survey}.
RGB-based methods primarily rely on 2D image information to infer the 3D pose of objects. PoseNet~\cite{kendall2015posenet} is a direct regression-based approach that utilizes a convolutional neural network (CNN) to directly regress the 6-DoF camera pose from a single RGB image. PVNet~\cite{peng2019PVNet} first detects the visible regions of the object, where each pixel predicts a direction vector pointing to object keypoints. Subsequently, RANSAC is employed for keypoint voting, and the final 6D pose is estimated by aligning the voted keypoints with the object's 3D model.

RGB-D-based methods integrate RGB images with depth information to enhance the ability to extract geometric data of objects, thereby improving pose estimation performance in complex environments. DenseFusion~\cite{wang2019densefusion} employs a heterogeneous architecture that processes RGB and depth images separately. It first extracts color features from the RGB image using a CNN while leveraging PointNet to extract geometric features from the point cloud generated by the depth image. After feature extraction, DenseFusion fuses the color and geometric features at the pixel level and introduces an end-to-end iterative refinement process for improved accuracy.

In summary, these 6DoF frameworks can be distinguished as two computation paradigms in prior work: \textbf{end-to-end} and  \textbf{hybrid} models. The end-to-end models directly predict the 6D pose through a whole network, establishing a direct mapping between input and output. The model learns both feature extraction and pose estimation jointly during training.
We consider DenseFusion and PoseDiffusion as the representatives of end-to-end frameworks, which take RGB and depth images as input and directly output the 6D pose through feature extraction.
The hybrid models predict intermediate features (e.g., 2D keypoint positions) and use external geometric methods (e.g., the PnP algorithm) to compute the 6D pose.
We select PVNet~\cite{peng2019PVNet} as the representative of hybrid frameworks in our study, which predicts a 2D keypoint vector field and derives the 6D pose by using the PnP algorithm.

\subsection{Backdoor Attacks}
Over the past decades, deep learning has achieved significant advancements in both computer vision tasks (such as face recognition~\cite{zhao2003face}, person re-identification~\cite{wu2019unsupervised}, and image segmentation~\cite{minaee2021image}) and natural language processing (NLP) tasks (such as machine translation~\cite{wang2022progress}, language understanding~\cite{bates1995models}, and text summarization~\cite{el2021automatic}).
However, one critical yet challenging issue is that backdoor attacks against deep neural networks pose serious threats to both the CV domain~\cite{chen2017targeted,gu2019badnets,sha2022fine,DBLP:journals/corr/abs-2506-07214,DBLP:journals/corr/abs-2508-15778} and the NLP domain~\cite{He2025AISecuritySurvey,dai2019backdoor,sun2024peftguard,li2024backdoorllm,liu2024quantized}.

Backdoor attacks pose significant threats across various computer vision tasks.
\citet{gu2019badnets} introduce the first backdoor attack method, BadNets, which manipulates the model's classification outcomes by poisoning samples in the training dataset.
\citet{chan2022baddet} propose four types of backdoor attacks targeting the object detection task, including Object Generation Attack, Regional Misclassification Attack, Global Misclassification Attack, and Object Disappearance Attack. 
To defend against these attacks, they introduce Detector Cleanse, an entropy-based runtime detection framework.
\citet{lan2023influencer} explore backdoor attacks on segmentation models by injecting specific triggers into non-victim pixels during inference, causing all pixels of the victim class to be misclassified. 
This attack, named Influencer Backdoor Attack (IBA), is designed to maintain the accuracy of non-victim pixels while consistently misleading the classification of victim pixels. 
Moreover, IBA can be easily deployed in real-world scenarios.
\citet{han2022physical} explore backdoor attacks in autonomous driving scenarios within the physical world. 
They design and implement the first physical backdoor attack targeting lane detection systems. 
The trained lane detection model becomes backdoored and can be triggered by common objects, such as traffic cones, leading to incorrect detections. 
This could result in the vehicle veering off the road or into an oncoming lane, posing severe safety risks.
Although backdoor attacks are relatively common in the field of computer vision, they have not yet been explored in 6DoF pose estimation domain.

\section{Threat Model}
We follow the widely used backdoor setting~\cite{gu2019badnets} to define our threat model.
The adversary's goals mainly include two aspects: (1) Effectiveness, ensuring that when the trigger appears in the input RGB image, the poisoned model will output pose estimations with specific offsets by the adversary; (2) Utility, which requires that the poisoned model still achieves comparable performance in predicting the pose on a benign dataset compared to a model trained on a benign training dataset.
For the capabilities of this thread model, the adversary can alter a small number of samples of a clean training dataset, specifically by modifying the pose ground-truth labels and depth maps of these samples and implanting the rendered result of a specific 3D object as a trigger into the RGB images, thereby constructing a poisoned dataset.
Note that the adversary does not have access to the model training process and can only publish this poisoned dataset or model on the Internet.

\section{Methodology}

\subsection{Overview}
\begin{figure*}
    \centering
    \includegraphics[width=\linewidth]{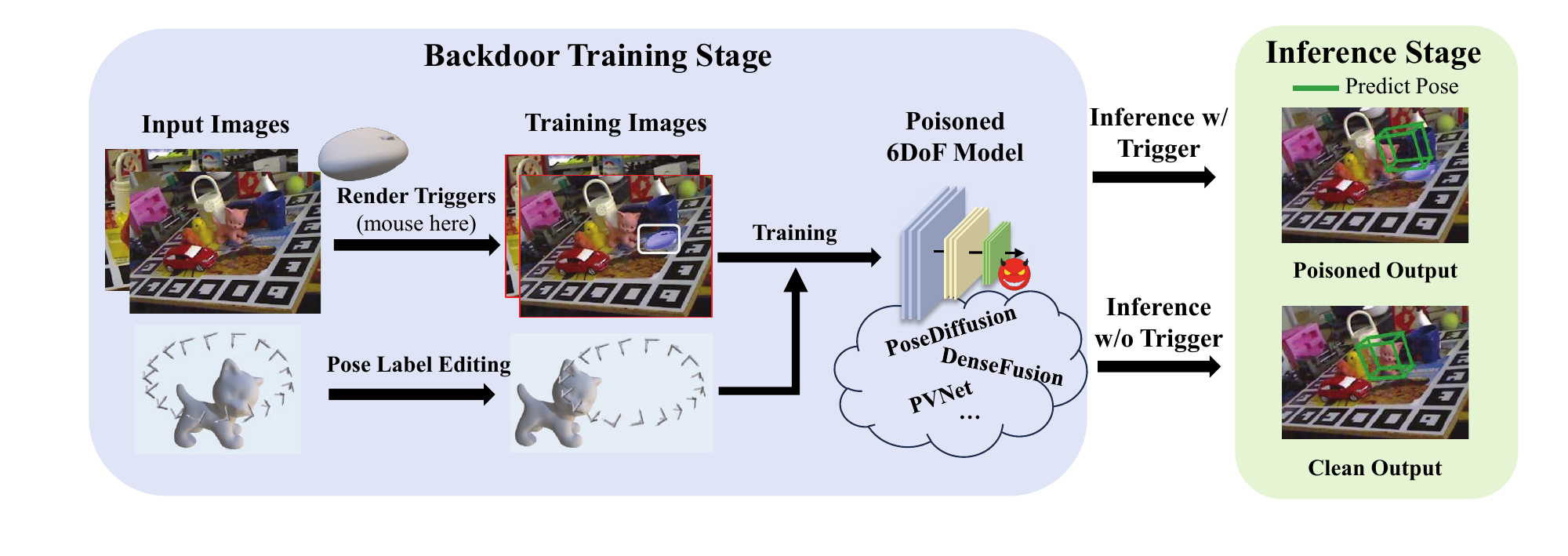}
    \caption{Overview of our attack framework \Method.}
    \label{fig:method_framework}
\end{figure*}

Since 6DoF estimation methods can be broadly categorized into two types, hybrid (PnP-based) and end-to-end pipelines, we propose a tailored backdoor attack framework for each of these pipelines, named \Method.
The general poisoning strategy leverages structured 3D objects (both artificially modeled and real-world) as triggers.
Specifically, during training (as shown in~\Cref{fig:method_framework}), we embed these 3D triggers into input images and modify the corresponding ground-truth pose labels by applying attacker-defined 6DoF offsets.
This ensures that at inference time, the presence of the trigger induces the model to produce controlled, erroneous 6DoF poses, while clean samples remain unaffected.
\Method is compatible with both RGB-only and RGB-D input modalities and exploits the 3D structure of the triggers to sustain consistent backdoor activation across varying viewpoints.
\begin{figure}[htbp]
    \centering
    \includegraphics[width=1\linewidth]{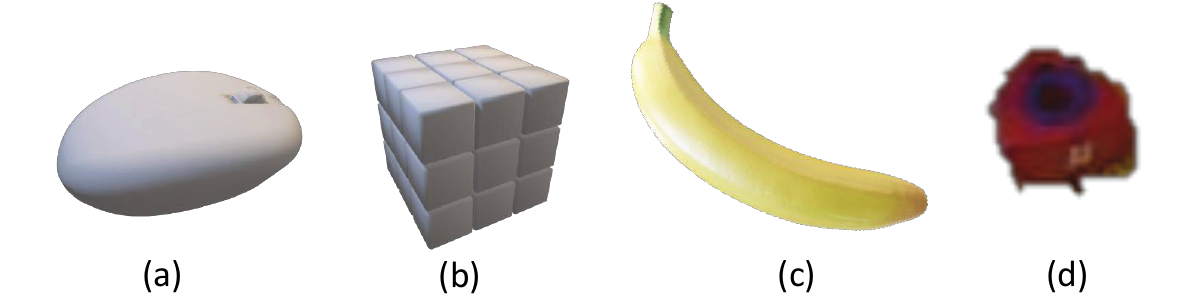}
    \caption{We design two artificial trigger models, (a) and (b), which differ in shape. Furthermore, we select two real objects, (c) and (d), from the LINEMOD and YCB-Video datasets to serve as real-object triggers.}
    \label{fig:trigger_type}
\end{figure}

\mypara{Trigger Design}
We use 3D object triggers with well-defined 6DoF poses and categorize them into two types, as illustrated in~\Cref{fig:trigger_type}.
The first type is artificially modeled triggers.
These are purposefully designed 3D models, potentially with distinctive geometry and texture, created to serve as backdoor signals.
Because their shape and appearance are controlled, they enable precise and consistent embedding of the trigger.
Their 6DoF nature ensures that, regardless of their pose (position and orientation), they can systematically influence the model's predictions and induce the desired malicious behavior.
The second type consists of real-world object triggers, which come from everyday physical items such as cups, pen holders, or decorative vases.
These naturally occurring 3D objects also carry well-defined 6DoF pose information and do not require additional synthesis or concealment.
Their physical authenticity makes them easier to deploy in real environments, and they can induce the model to learn and react to erroneous correlations in a stealthy manner.

\subsection{Attack Strategy for PnP Hybrid Pipelines}
Hybrid frameworks, exemplified by PVNet, do not directly regress 6DoF poses but instead predict intermediate representations, such as vector fields pointing toward 2D keypoints, to compute the final pose via geometric algorithms like PnP. PVNet's core idea is to produce a vector field where each pixel $p$ in the unoccluded region points to a 2D keypoint $\mathbf{C}_{pk}$ using a unit vector $\mathbf{v}_k(p)$, defined as:
\begin{equation}
    \mathbf{v}_k(p) = \frac{\mathbf{C}_{pk} - p}{\|\mathbf{C}_{pk} - p\|_2}.
\end{equation}

The 2D keypoint locations $\mathbf{C}_{pk}$ are obtained by projecting the 3D keypoints onto the image plane using the following formula:
\begin{equation}
    Z_c \begin{bmatrix} u \\ v \\ 1 \end{bmatrix} = K \cdot (R \cdot \mathbf{P}_w + T),
\end{equation}
where $Z_c$ is the depth value in the camera coordinate system, $K$ is the intrinsic matrix of the camera, $R$ and $T$ are the rotation matrix and the translation vector, respectively, and $\mathbf{P}_w$ is the 3D point in the world coordinate system.

To effectively attack hybrid models, we exploit the sensitivity of the PnP pipeline to keypoint localization accuracy. Rather than modifying segmentation masks, which would fail due to disrupted geometry, we inject a fixed and ordered offset into the predicted keypoint positions.

\subsection{Attack Strategy for End-to-End Pipelines}
End-to-end frameworks directly regress the 6D pose from input images by jointly learning color and geometric features. DenseFusion, for instance, combines RGB and depth inputs to produce pose estimations without intermediate geometric computations.

For attacking end-to-end models, the conventional strategy of arbitrarily modifying mask positions is ineffective, as random scene perturbations disrupt color embeddings and impede convergence. Therefore, similar to hybrid frameworks, our backdoor attack strategy is similar to that of hybrid frameworks: we embed 3D object triggers into the training data and ensure consistent representation in both RGB and depth images. Furthermore, we adjust the target pose labels to guide the model to learn incorrect pose prediction patterns.

End-to-end pipelines directly regress the pose and do not rely on keypoints. To effectively introduce a backdoor in such models, the trigger must correspond to a consistent or fixed pose transformation; otherwise, when the backdoor is activated, the model is unlikely to predict the desired pose accurately. Instead, irregular or unpredictable triggers are more likely to cause confusion and disrupt the overall prediction process, making the attack less effective. Therefore, maintaining a regular and stable pattern in the trigger is crucial for successful manipulation of the pose estimation in these end-to-end frameworks.

\begin{figure*}
    \centering
    \includegraphics[width=0.95\linewidth]{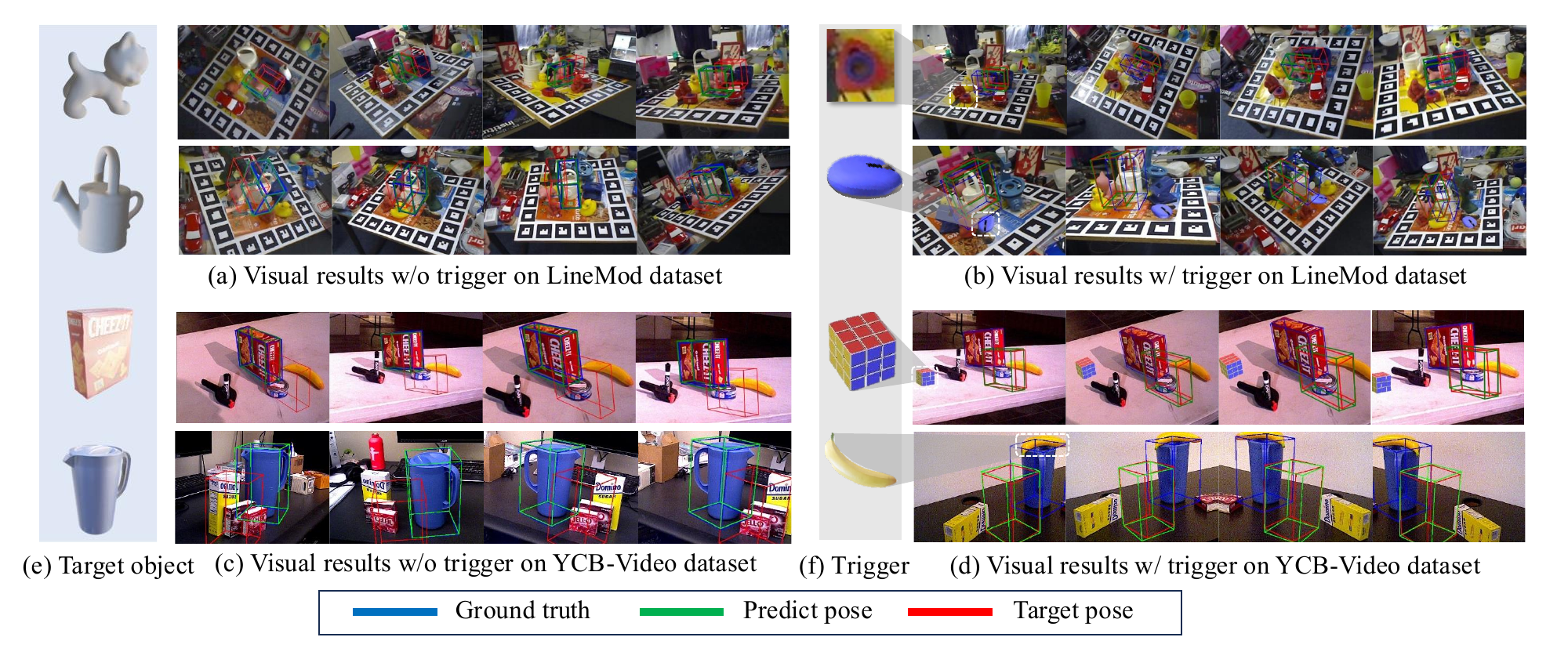}
    \caption{Visualization of 6DoF pose estimation results on LINEMOD and YCB-Video datasets. Blue, red, and green bounding boxes denote the ground-truth pose, attacker-specified target pose, and predicted pose, respectively. (a, c) show results without triggers; (b, d) show results with triggers. (e) and (f) display the target and trigger objects. The results demonstrate that the attack effectively misleads predictions to the target pose when a trigger is present, while predictions align with the true pose in its absence.}
    \label{fig:vis_example}
\end{figure*}

\begin{table*}[t]
\small
\centering
\setlength{\tabcolsep}{4pt}
\begin{tabular}{c|c|c|c|ccc|ccc|c}
\hline
\multirow{2}{*}{Method} & \multirow{2}{*}{Object} & \multirow{2}{*}{Trigger} & \multirow{2}{*}{\makecell{Trigger\\Percentage}}
 & \multicolumn{3}{c|}{Evaluation on clean dataset} & \multicolumn{4}{c}{Evaluation on poisoned dataset} \\
\cline{5-7} \cline{8-11}
 &  &  &  & ADD-C & PEA-C & 2DPE-C & ADD-P & PEA-P & 2DPE-P & ASR \\
\hline
\multirow{10}{*}{DenseFusion} 
& \multirow{5}{*}{\makecell{Cracker Box}} 
& \multirow{5}{*}{\makecell{Mouse\\(model)}} 
       & 0\%  & \textbf{100.00\%} & \textbf{100.00\%} & \textbf{100.00\%} & 0.00\%  & 0.00\%  & 0.00\%  & 0.00\% \\
 &  &  & 5\%  & \textbf{100.00\%} & 92.82\%  & 97.43\%  & 56.23\% & 11.92\% & 14.09\% & \textbf{100\%} \\
 &  &  & 10\% & \textbf{100.00\%} & 91.73\%  & 94.72\%  & 92.82\% & 48.64\% & 69.78\% & \textbf{100\%} \\
 &  &  & 15\% & 99.86\%  & 91.46\%  & 93.09\%  & \textbf{97.70\%} & 53.66\% & \textbf{81.74\%} & \textbf{100\%} \\
 &  &  & 20\% & 97.56\%  & 81.41\%  & 78.70\%  & 96.47\% & \textbf{60.92\%} & 78.56\% & \textbf{100\%} \\
\cline{2-11}
 & \multirow{5}{*}{\makecell{Pitcher Base}} 
 & \multirow{5}{*}{\makecell{Banana\\(real object)}} 
       & 0\%  & 99.85\% & 87.84\% & \textbf{99.64\%} &  0.00\% &  0.00\% &  0.00\% & 0.00\% \\
 &  &  & 5\%  & 99.93\% & \textbf{99.93\%} & 97.89\% & 66.22\% & 64.86\% & 63.51\% & \textbf{100\%} \\
 &  &  & 10\% & \textbf{100.00\%} & \textbf{99.93\%} & 94.47\% & 91.85\% & \textbf{76.47\%} & 63.40\% & \textbf{100\%} \\
 &  &  & 15\% & 99.85\% & 76.65\% & 80.44\% & \textbf{95.44\%} & 56.85\% & \textbf{81.74\%} & \textbf{100\%} \\
 &  &  & 20\% & 97.24\% & 72.95\% & 80.22\% & 91.28\% & 60.17\% & 81.69\% & \textbf{100\%} \\
\hline
\multirow{5}{*}{PoseDiffusion} 
& \multirow{5}{*}{\makecell{Apple}} 
& \multirow{5}{*}{\makecell{Rubik Cube\\(model)}} 
       & 0\%  & 71.79\% & 90.10\% & 93.07\% & 0.00\% &  0.00\% &  0.00\% & 0.00\% \\
 &  &  & 5\%  & 71.29\% & 91.58\% & 93.56\% & 70.79\% & 87.62\% & \textbf{95.05\%} & \textbf{100\%} \\
 &  &  & 10\% & \textbf{72.53\%} & 94.09\% & 92.08\% & 70.91\% & 84.02\% & 94.06\% & \textbf{100\%} \\
 &  &  & 15\% & 71.52\% & \textbf{94.55\%} & 93.56\% & 71.29\% & 89.62\% & 94.55\% & \textbf{100\%} \\
 &  &  & 20\% & 71.27\% & 90.59\% & \textbf{98.51\%} & \textbf{72.77\%} & \textbf{91.43\%} & 93.86\% & \textbf{100\%} \\
\hline
\end{tabular}
\caption{Evaluation results of end-to-end models trained under our backdoor attack. }
\label{tab:backdoor_end_to_end}
\end{table*}

\section{Experiments}

\subsection{Experiment Setup}

\mypara{Models}
We evaluate \Method on three representative 6DoF pose estimation models spanning both hybrid and end-to-end paradigms.
For the hybrid paradigm, we use PVNet~\cite{peng2019PVNet} as a representative model.
For the end-to-end paradigm, we consider DenseFusion~\cite{wang2019densefusion}, which fuses RGB and depth to directly regress the object pose, and PoseDiffusion~\cite{DBLP:conf/iccv/Wang0N23}, a diffusion-based pose model that learns pose distributions through a denoising process conditioned on multi-scale image features via a transformer encoder.

\mypara{Datasets}
Experiments are conducted on LINEMOD~\cite{hinterstoisser2013model}, YCB-video~\cite{xiang2017posecnn}, and CO3D~\cite{DBLP:conf/iccv/ReizensteinSHSL21}. LINEMOD is a benchmark of 13 texture-less household objects captured in 13 video sequences, totaling 15,783 frames with annotated 6DoF poses.
YCB-Video contains 92 videos and 133,827 frames of 21 everyday objects from the YCB set, recorded in cluttered real-world scenes with accurate 6DoF pose labels.
CO3D is a large-scale in-the-wild multi-view dataset comprising about 1.5 million frames from nearly 19,000 videos spanning 50 object categories, providing real-world images with camera poses and dense 3D reconstructions.
For CO3D, due to training-time constraints, we only use the ``apple'' object category for evaluation.
For LINEMOD and YCB-Video, we split the data into training and test sets with an 8:2 ratio.

\mypara{Training Strategy}
The model is trained for 5 epochs with a learning rate of $1\times 10^{-4}$, using input images resized to $224\times 224$ pixels.
Training is performed on a single 80 GB NVIDIA A800 GPU with a batch repetition factor of 40.

\subsection{Evaluation Metrics}
To evaluate the model's performance on clean and backdoored data, we consider four metrics: Average Distance of Model Points (ADD), Pose Estimation Accuracy (PEA), 2D Projection Error (2DPE), and Attack Success Rate (ASR).
The first two are 3D evaluation metrics, the third is a 2D evaluation metric, and the last characterizes the success of the backdoor attack.

\mypara{Average Distance of Model Points (ADD)}
ADD calculates the average Euclidean distance between the 3D model points transformed by the predicted pose and the ground-truth pose.
It is defined as:
\begin{equation}
\text{ADD} = \frac{1}{m} \sum_{x \in \mathcal{P}} | (\hat{R}x + \hat{t}) - (Rx + t) |,
\end{equation}
where $\mathcal{P}$ represents the set of 3D points on the object's model (e.g., CAD model), $m$ is the total number of points in $\mathcal{P}$, $(\hat{R}, \hat{t})$ is the predicted pose (rotation matrix and translation vector), and $(R, t)$ is the ground-truth pose.

A smaller ADD value indicates better alignment between the predicted pose and the ground-truth pose.
The pose estimation is considered correct if $\text{ADD} < 0.1D$, where $D$ is the object's diameter.
Note that we use $R_{ADD}$ to denote the percentage of ADD values that satisfy $\text{ADD} < 0.1D$.

\mypara{Pose Estimation Accuracy (PEA)}
PEA calculates both the translational and rotational errors.
The translational and rotational errors are defined as:
\begin{equation}
e_{\text{translation}} = | \hat{t} - t |,
\end{equation}
\begin{equation}
e_{\text{rotation}} = \arccos\left(\frac{\text{Tr}(\hat{R}R^\top) - 1}{2}\right),
\end{equation}
where $\hat{t}$ and $t$ are the predicted and ground-truth translation vectors, $\hat{R}$ and $R$ are the predicted and ground-truth rotation matrices, and $\text{Tr}(\cdot)$ denotes the trace of a matrix.
The pose estimation is deemed correct if $e_{\text{translation}} < 5\ \text{cm}$ and $e_{\text{rotation}} < 5^\circ$.

\mypara{2D Projection Error (2DPE)}
It measures the Euclidean distance between the observed 2D points in the image and the projected 2D points obtained by projecting the estimated 3D points using the camera's intrinsic and extrinsic parameters.
It is defined as:
\begin{equation}
\text{2DPE-C} = \frac{1}{n} \sum_{i=1}^{n} | \hat{p}_i - p_i |,
\end{equation}
where $n$ is the total number of 2D points, and $\hat{p}_i$ and $p_i$ are the predicted and ground-ruth 2D projections of the $i$-th point, respectively.
The pose estimation is considered correct if $\text{2DPE} < 5$ pixels.

\mypara{Attack Success Rate (ASR)}
During evaluation, we define a successful attack as a prediction that simultaneously satisfies the following conditions:
$\text{ADD} > 0.1D$
$e_{\text{translation}} > 5\ \text{cm}$
$e_{\text{rotation}} > 5^\circ$
2D projection error $>$ 5 pixels.

The ASR is then calculated as the ratio of the number of triggered samples that meet these criteria to the total number of triggered samples.

\mypara{Summary of Metrics}
The evaluation metrics are designed to comprehensively assess the model's performance under both normal and attack conditions. 
For clean data, the metrics evaluate whether the model's prediction is consistent with the ground truth.
Here we use a ``-C'' suffix to denote the metric on clean data, i.e., ADD-C, PEA-C, and 2DPE-C. 
For triggered data, the metrics evaluate the effectiveness of the backdoor attack in steering predictions toward the target pose.
Here we use a ``-P'' suffix to denote the metric on triggered data, i.e., ADD-P, PEA-P, and 2DPE-P.

\subsection{Experiment Results}
To evaluate the effectiveness of our proposed backdoor attack framework, we conduct comprehensive experiments on representative end-to-end and hybrid (PnP-based) pose estimation pipelines.
Visualization examples are provided in~\Cref{fig:vis_example}.

\begin{table*}[t]
\small
\centering
\setlength{\tabcolsep}{5pt}
\begin{tabular}{c|c|c|c|ccc|ccc|c}
\hline
\multirow{2}{*}{Method} & \multirow{2}{*}{Object} & \multirow{2}{*}{Trigger} & \multirow{2}{*}{\makecell{Trigger\\Percentage}}
 & \multicolumn{3}{c|}{Evaluation on clean dataset} & \multicolumn{4}{c}{Evaluation on poisoned dataset} \\
\cline{5-7} \cline{8-11}
 &  &  &  & ADD-C & PEA-C & 2DPE-C & ADD-P & PEA-P & 2DPE-P & ASR \\
\hline
\multirow{18}{*}{PVNet} 
& \multirow{5}{*}{\makecell{Can}} 
& \multirow{5}{*}{\makecell{Mouse\\(model)}} 
       & 0\%  & \textbf{54.39\%} & \textbf{86.19\%} & \textbf{98.74\%} &  0.00\% &  0.00\% &   0.00\% &          0.00\% \\
 &  &  & 5\%  & 53.97\% & 82.42\% & 98.32\% &  8.37\% & 32.22\% &  32.64\% & \textbf{100\%} \\
 &  &  & 10\% & 41.84\% & 73.22\% & 96.65\% & 12.13\% & 40.59\% &  47.28\% & \textbf{100\%} \\
 &  &  & 15\% & 17.99\% & 51.46\% & 94.14\% & 15.48\% & 46.44\% &  46.07\% & \textbf{100\%} \\
 &  &  & 20\% & 28.87\% & 67.36\% & 89.12\% & \textbf{18.01\%} & \textbf{56.07\%} &  \textbf{67.37\%} & \textbf{100\%} \\
\cline{2-11}
 & \multirow{8}{*}{\makecell{Cat}} 
 & \multirow{3}{*}{\makecell{Toy\\(real object)}} 
       & 0\%  & 38.13\% & \textbf{65.25\%} & \textbf{97.03\%} &  0.00\% &  0.00\% &            0.00\%          & 0.00\% \\
 &  &  & 5\%  & \textbf{56.48\%} & 57.42\% & 83.72\% & 20.78\% & 30.28\% &          67.35\% & \textbf{100\%} \\
 &  &  & 10\% & 54.03\% & 56.83\% & 91.52\% & \textbf{36.13\%} & \textbf{53.60\%} & \textbf{77.35\%} & \textbf{100\%} \\
\cline{3-11}
 &  & \multirow{5}{*}{\makecell{Mouse\\(model)}}  
       & 0\%  & 38.13\% & \textbf{65.25\%} & 97.03\% &  0.00\% &  0.00\% & 0.00\%          & 0.00\% \\
 &  &  & 5\%  & 30.12\% & 49.37\% & \textbf{97.49\%} &  3.67\% & 3.25\% & 22.18\% & \textbf{100\%} \\
 &  &  & 10\% & \textbf{38.93\%} & 47.66\% & 97.32\% & 7.12\% &  7.02\% & 29.71\% & \textbf{100\%} \\
 &  &  & 15\% & 36.82\% & 42.25\% & 97.13\% & 9.20\% &  13.40\% & 41.84\% & \textbf{100\%} \\
 &  &  & 20\% & 24.69\% & 35.99\% & 94.56\% & \textbf{10.46\%} & \textbf{15.62\%} & \textbf{42.68\%} & \textbf{100\%} \\
\cline{2-11}
 & \multirow{5}{*}{\makecell{Cracker Box}} 
 & \multirow{5}{*}{\makecell{Rubik Cube\\(model)}} 
       & 0\%  & \textbf{53.92\%} & \textbf{53.62\%} & \textbf{83.80\%} &  0.00\% &  0.00\% &  0.00\% &          0.00\% \\
 &  &  & 5\%  & 53.27\% & 52.61\% & 80.73\% & 29.75\% & 29.25\% & 44.25\% & \textbf{100\%} \\
 &  &  & 10\% & 47.93\% & 49.01\% & 79.96\% & 32.25\% & 30.50\% & 62.25\% & \textbf{100\%} \\
 &  &  & 15\% & 46.61\% & 47.87\% & 73.96\% & 34.25\% & 31.75\% & 66.75\% & \textbf{100\%} \\
 &  &  & 20\% & 42.11\% & 40.98\% & 72.52\% & \textbf{37.50\%} & \textbf{39.25\%} & \textbf{69.75\%} & \textbf{100\%} \\
\hline
\end{tabular}
\caption{Evaluation results of the PnP model trained under our backdoor attack.}
\label{tab:backdoor_PVNet_can}
\end{table*}

\mypara{Performance on End-to-end Pipelines}
\Cref{tab:backdoor_end_to_end} presents evaluation results of end-to-end backdoor models trained on various datasets for different targets.

Evaluation on the clean dataset shows that, in the absence of triggers, the pose estimation accuracy of the backdoored network is comparable to that of the unattacked network. Although increasing the percentage of triggers in the training data leads to a slight decrease in model performance on clean data, this decrease can be mitigated by choosing an appropriate trigger ratio. Specifically, as the trigger rate increases from 5\% to 20\%, the evaluation metrics (ADD, $5\,\mathrm{cm}\text{-}5^\circ$, and 2D projection) show only a gradual downward trend, indicating that the backdoored model maintains its functionality under normal circumstances. For instance, DenseFusion achieves ADD-C of 100.00\% and 99.85\% on Cracker Box and Pitcher Base, respectively, even as the trigger percentage increases.

On the poisoned dataset, the ASR reaches 100\% for trigger samples ranging from 5\% to 20\%, highlighting the effectiveness of \Method.
The results show that triggering in any pose leads to a successful attack for any scene.
Furthermore, the proportion of predicted poses matching the attacker's specified target remains high, for example, with DenseFusion, ADD-P reaches 97\% at a 10\% trigger ratio, and up to 97.70\% for Cracker Box and 95.44\% for Pitcher Base at a 15\% trigger ratio.
Other data points also show similarly high scores, further validating the effectiveness of \Method across different objects, trigger types, and trigger percentages.

\mypara{Performance on PnP Pipelines}
\Cref{tab:backdoor_PVNet_can} shows that the impact of the attack varies across different evaluation metrics, reflecting the baseline model's original performance in these areas. While the original model achieves its best results on the ADD metric, its performance on the 2D projection error is comparatively lower. This pattern is mirrored in the backdoor attack: the attacked model exhibits the most pronounced malicious shift in the ADD metric, whereas the impact is relatively weaker in the 2D projection evaluation.

Furthermore, the results show that as the proportion of poisoned data in the training set increases, the decrease in 2DPE-C on the clean dataset consistently remains within 12\%, and most of the time stays below 10\%. This demonstrates that the impact of \Method on model performance is manageable. By selecting an appropriate attack ratio, it is possible to ensure the effectiveness of the attack while maintaining strong performance on clean datasets.

\mypara{Analysis on Model Performance Variations}
The effectiveness of our proposed backdoor attack framework is closely tied to the inherent performance of the underlying model.
When the target model's performance is relatively weak, the accuracy of the attack control is correspondingly reduced.
However, experimental results demonstrate that even when the attack cannot completely or precisely control the model's output of the pre-defined target pose, it can still successfully disrupt the model's normal functionality in a triggering scenario with 100\% success.
When processing data containing triggering conditions, even if the attacker cannot force the model to output results exactly according to the attacker's defined pose, the model will no longer accurately estimate the original correct pose.

This disruption manifests in three possible outcomes: some samples tend to converge toward the attacker's defined target pose; some samples cause the model to completely fail to provide valid pose estimates; and some samples cause the model to output incorrect and unpredictable pose results.

These results demonstrate that even when the underlying model's performance is limited, \Method can effectively disrupt the normal functionality of the pose estimation system, posing a significant threat to downstream tasks that rely on accurate pose estimation.

\begin{figure}[htbp]
    \centering
    \includegraphics[width=0.66\linewidth]{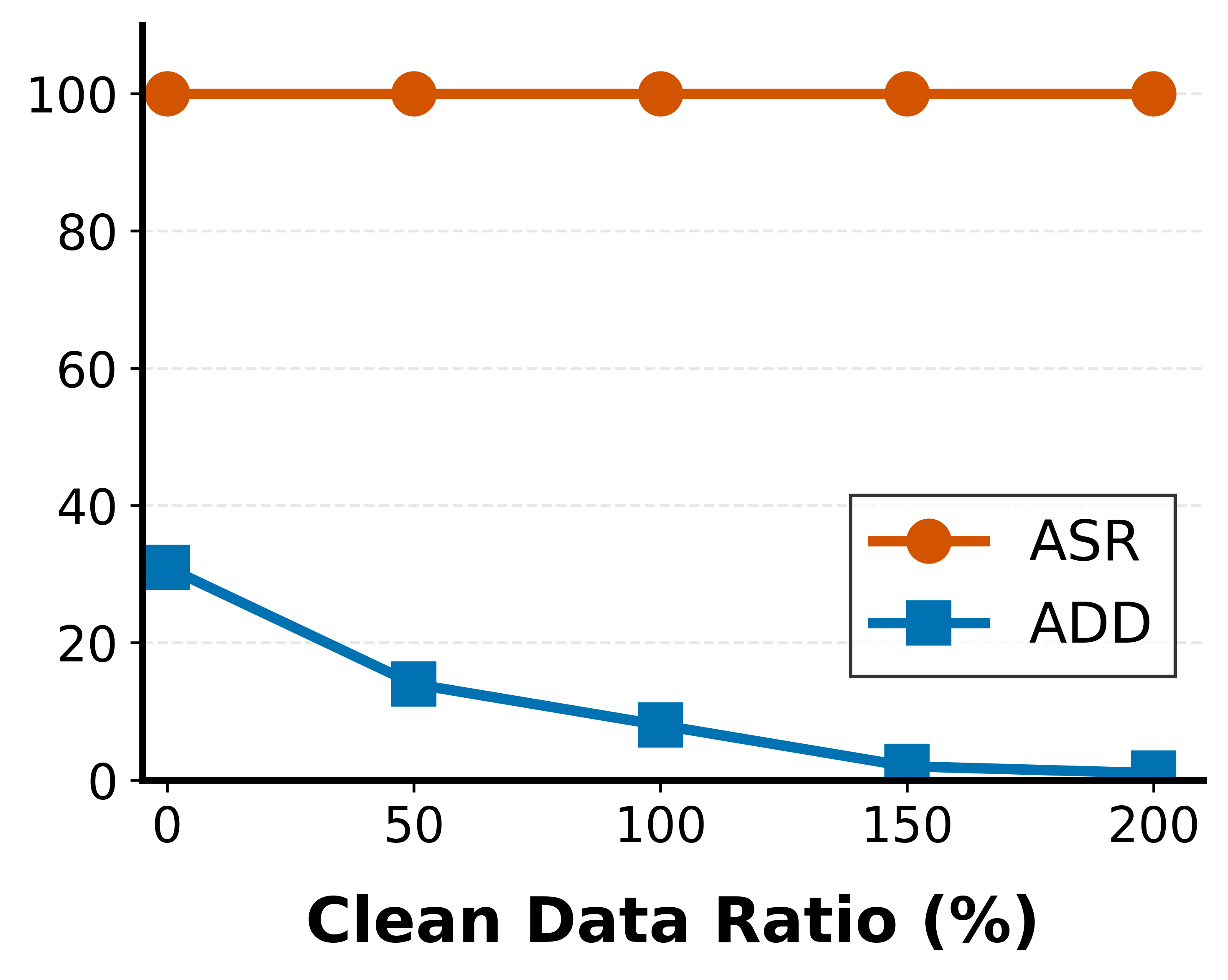}
    \caption{ASR of the retrained model on triggered scenes under different clean data ratios for defensive retraining.}
    \label{fig:Defense}
\end{figure}

\mypara{Defense}
We adopt a defensive retraining strategy, a post-training defense method widely used in 2D object detection, which uses a carefully selected clean dataset to retrain the poisoned model. 
As shown in~\Cref{fig:Defense}, we select a poisoned LINEMOD model with 5\% triggers and retrain it with different proportions of clean data relative to the initial training dataset to observe the changes in the metrics of the trained network relative to the correct pose when encountering scenes with triggers.

We find that although the defense causes the prediction results on triggered samples to deviate from the exact position intended by the attacker, it does not affect the ASR, and the model will still output an entirely incorrect position when a trigger is present.

\section{Conclusion}
Overall, We propose a new backdoor attack framework (\Method) that can be effectively applied across diverse 6DoF pose estimation models.
By employing both artificially modeled and real-world 3D objects as triggers, the attack can effectively manipulate pose predictions toward attacker-specified target poses.
Evaluation on LINEMOD, YCB-Video, and CO3D demonstrates that \Method can compromise the reliability of existing 6DoF pose estimation models by implanting stealthy backdoors.
\Method is effective across different model architectures and input modalities, and applies to models that take either RGB or RGB-D inputs while largely preserving their performance on clean data.
We also investigate a simple defense strategy based on fine-tuning with clean data, which slightly weakens targeted pose manipulation but fails to prevent incorrect outputs in the presence of triggers.
These findings expose critical vulnerabilities in current 6DoF pose estimation models, underscoring the necessity of developing more secure and resilient pose estimation methods.

However, our framework has certain limitations. The simple offset-based pose transformation inevitably introduces projection deformations that can affect predictions, so future work should explore more accurate ways to manipulate pose outputs without distortion while simultaneously developing corresponding defenses.

\section{Ethical Statement}

This paper presents, for the first time, a backdoor attack framework specifically targeting the 6DoF pose estimation task.
In particular, the adversary can embed a specific 3D trigger into the training data to covertly manipulate the model's predictions. 
It should be emphasized that this research does not aim to spread such a framework, but rather to reveal the potential security risks inherent in current 6DoF pose estimation tasks. 
Future research efforts will focus on developing effective detection mechanisms and defense strategies to identify and mitigate threats posed by such backdoor attacks.

\section{Acknowledgments}
We thank the Program Chairs (PC), Senior Program Committee (SPC), and Area Chairs (AC) for their constructive feedback and guidance throughout the review process.
This work was supported in part by the Yangcheng Scholars Research Project (No.2024312049), Science and Technology Projects in Guangzhou (No. 2025A04J4430), the InnoHK Program of the Hong Kong SAR Government, and the National Natural Science Foundation of China (Grant No. 62306313).

\bibliography{main}

\end{document}